\documentclass{article}

\usepackage[nonatbib,final]{neurips_2023}
\usepackage{graphicx}

\usepackage[pagebackref=true,breaklinks=true,letterpaper=true,colorlinks,bookmarks=false, citecolor=ForestGreen]{hyperref}
\hypersetup{
    colorlinks=true,
    urlcolor=magenta 
}
\usepackage[dvipsnames,table,xcdraw]{xcolor}

\usepackage[utf8]{inputenc} 
\usepackage[T1]{fontenc}    
\usepackage{url}            
\usepackage{booktabs}       
\usepackage{amsfonts}       
\usepackage{nicefrac}       
\usepackage{microtype}      
\usepackage{tabularx}
\usepackage{subfigure}

\newcommand{\red}[1]{{\color{red}{#1}}}

\newcommand{\B}{\mathcal{B}}
\newcommand{\C}{\mathcal{C}}

\newcommand{\I}{\mathcal{I}}

\newcommand{\X}{\mathcal{X}}
\newcommand{\R}{\mathcal{R}}

\newcommand{\M}{\mathcal{M}}

\newcommand{\qileft}{[\kern-0.15em[}
\newcommand{\qiLeft}{\left[\kern-0.32em\left[}
\newcommand{\qiright}{]\kern-0.15em]}
\newcommand{\qiRight}{\right]\kern-0.32em\right]}

\newcommand{\qiLeftm}{\left[\kern-0.28em\left[}
\newcommand{\qiRightm}{\right]\kern-0.28em\right]}

\newcommand{\qiLefts}{\left[\kern-0.16em\left[}
\newcommand{\qiRights}{\right]\kern-0.16em\right]}

\title{Detecting Any Human-Object Interaction Relationship: Universal HOI Detector with Spatial Prompt Learning on Foundation Models}

\author{
Yichao Cao\\
Southeast University\\
{\tt\small }
\and
Qingfei Tang\\
Nanjing Enbo Tech.\\
{\tt\small }
\and
Xiu Su\thanks{Corresponding author (xisu5992@uni.sydney.edu.au). First author (caoyichao@seu.edu.cn).}\\
University of Sydney\\
{\tt\small}
\and
Chen Song\\
Nanjing Enbo Tech.\\
{\tt\small}
\and
Shan You\\
SenseTime Research\\
{\tt\small }
\and
Xiaobo Lu\\
Southeast University\\
{\tt\small }
\and
Chang Xu\\
University of Sydney\\
{\tt\small }
}

\begin{document}

\maketitle

\begin{abstract}
Human-object interaction (HOI) detection aims to comprehend the intricate relationships between humans and objects, predicting $<human, action, object>$ triplets, and serving as the foundation for numerous computer vision tasks. The complexity and diversity of human-object interactions in the real world, however, pose significant challenges for both annotation and recognition, particularly in recognizing interactions within an open world context. This study explores the universal interaction recognition in an open-world setting through the use of Vision-Language (VL) foundation models and large language models (LLMs). The proposed method is dubbed as \emph{\textbf{UniHOI}}. We conduct a deep analysis of the three hierarchical features inherent in visual HOI detectors and propose a method for high-level relation extraction aimed at VL foundation models, which we call HO prompt-based learning. Our design includes an HO Prompt-guided Decoder (HOPD), facilitates the association of high-level relation representations in the foundation model with various HO pairs within the image. Furthermore, we utilize a LLM (\emph{i.e.} GPT) for interaction interpretation, generating a richer linguistic understanding for complex HOIs. For open-category interaction recognition, our method supports either of two input types: interaction phrase or interpretive sentence.  Our efficient architecture design and learning methods effectively unleash the potential of the VL foundation models and LLMs, allowing UniHOI to surpass all existing methods with a substantial margin, under both supervised and zero-shot settings. The code and pre-trained weights are available at: \url{https://github.com/Caoyichao/UniHOI}.
\end{abstract}

\section{Introduction}
Human-object interaction (HOI) detection \cite{gupta2015visual, chao2018learning} is a burgeoning field of research in recent years, which stems from object detection and makes higher demands on high-level visual understanding. It plays a crucial role in many vision tasks, e.g., visual question answering, human-centric understanding, image generation, and activity recognition, to name a few representative ones. An excellent HOI detector must accurately localize all the interacting Human-Object (HO) pairs and recognize their interaction, typically represented as an HOI triplet in the format of $<human, action, object>$.

Despite the optimistic progress made by various HOI detectors \cite{li2019transferable, gao2020drg, hou2021detecting, zhang2022efficient, gkioxari2018detecting, liao2020ppdm, chen2021reformulating} in recent years, numerous challenges still persist \cite{yuan2022rlip}. Naively mapping diverse interactions to one-hot labels is an exceedingly labor-intensive and costly process, which is prone to the loss of semantic information. As the number and complexity of interaction categories increase, model optimization becomes increasingly difficult, and the universality of the model is consequently restricted.

Recently, multi-modal learning, especially Vision-and-Language (VL) learning, has become remarkably popular, achieving far-reaching success across a variety of domains \cite{alayrac2022flamingo}. Multi-modal learning is capable of portraying multidimensional information in a more comprehensive manner by describing the same entity or spatio-temporal event through different modalities. Recently, a number of studies have endeavored to employ VL models for the HOI task with the aim of constructing transferable HOI detectors \cite{zhong2021polysemy, iftekhar2022look, li2022improving, wang2022learning, yuan2022rlip, cao2023re}. For instance, PhraseHOI \cite{li2022improving} harnessed a pre-trained word embedding model to generate a fixed-length linguistic representation for the HOI task. Unfortunately, these methods harness cross-modal knowledge in an overly constrained manner, falling short of fully exploiting the potential of cross-modal knowledge and LLMs within the HOI detection domain.

Upon rigorous examination, we identify several intrinsic limitations in the existing methodologies: (1) Limited scalability: Model training is overly reliant on annotated data, which restricts them to a finite set of categories. (2) Suboptimal adaptability in zero-shot settings: Even HOI-VLM approaches can only harness a limited number of word embeddings to accommodate unseen categories, thereby curtailing their adaptability. (3) Inability to quickly comprehend complex interactive behaviors from succinct descriptions, as humans can, indicating a dearth of a more universal and flexible framework for HOI detection.

\begin{figure*}[t]
\vspace{-7pt}
  \centering
  \includegraphics[width=\linewidth]{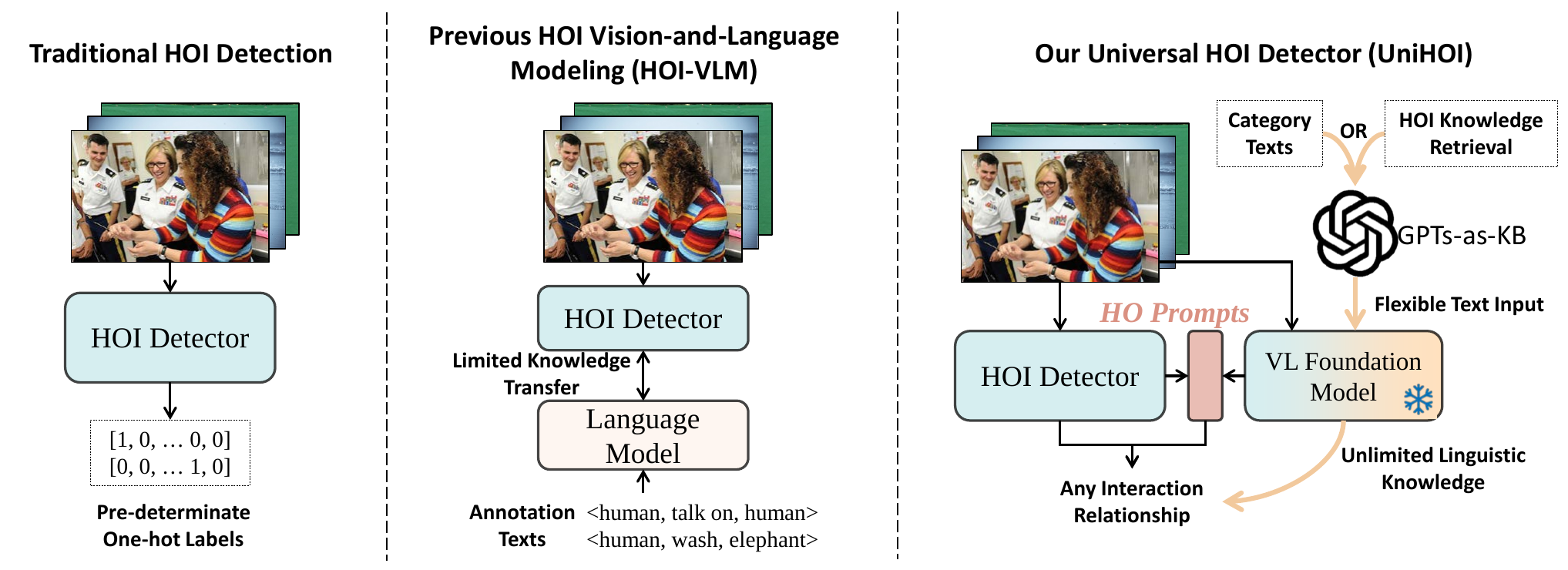}
    \vspace{-6pt}
   \caption{Comparison between HOI detection pipelines. Traditional HOI Detectors rely on manually annotated image datasets for training, which limits their generality and scalability. Previous HOI-VLM methods only achieved limited knowledge transfer from language models to visual HOI detectors, utilizing only a small amount of word embeddings for learning. The proposed UniHOI supports the input of any textual information (annotations or explanations) to detect any interaction relationship, fully unleashing the potential of foundation models and LLMs.}
   \label{fig:1}
\vspace{-16pt}
\end{figure*}

Considering the recent outstanding performance of VL foundation models and LLMs in various fundamental tasks, in this paper, we attempt to harness the capabilities of such large models to facilitate more general HOI detector, as shown in Figure \ref{fig:1}. We conduct an in-depth analysis of the clear three-tier visual feature hierarchy in visual HOI detectors, effectively utilizing the foundation models to boost the understanding of high-level interactive semantics within images. The employment of large models bridges the gap between textual and visual modalities, making a knowledge-based universal HOI detector feasible. In summary, the contributions of this paper are as follows:
\begin{itemize}
\item We propose the first visual-textual foundation model-based framework for HOI detection (UniHOI), which significantly improves the accuracy and universality compared to previous HOI detectors that solely rely on training from specific datasets. Our UniHOI method is expected to guide the HOI detection field into a new research phase.
\item We treat the visual HOI detector as a three-tier visual feature hierarchy and appropriately design an efficient human-object prompt learning method, enabling visual HOI detectors to effectively query interaction tokens associated with specific HO pairs in foundation models, and thus obtain a more universal HO relationship  representations.
\item We devise a knowledge-based zero-shot HOI recognition method that leverages large models such as GPT to generate descriptive knowledge for specific interaction categories, guiding visual detectors in zero-shot recognition of the corresponding categories. This approach breaks free from the limitations of traditional zero-shot recognition methods, which can only learn from word embeddings, and offers insights for a more universal open-category HOI recognition system.
\end{itemize}

\section{Related Work}

\textbf{Generic HOI Detection.} Based on their architectural design, existing HOI (Human-Object Interaction) detection approaches can be broadly categorized into two groups: two-stage methods \cite{zhang2022efficient, chao2018learning, gao2020drg, gao2018ican, li2019transferable, hou2021detecting}, and one-stage methods \cite{gkioxari2018detecting, kim2020uniondet, zhong2021glance, liao2020ppdm, chen2021reformulating}. One-stage methods typically adopt a multitask learning strategy to simultaneously execute instance detection and interaction relation modeling \cite{liao2020ppdm, zhang2021mining, liao2022gen}. Conversely, two-stage methods initially conduct object detection, followed by interaction relation modeling for all candidate HO pairs. By fully exploiting the capabilities of each module, two-stage methods have showcased superior detection performance \cite{zhang2021mining}. While these methods have significantly propelled the progress of early HOI detection, they still grapple with the issue of limited generality.

\textbf{Language Semantics for Vision.} Driven by the notable success of large-scale models such as GPT-4 \cite{openai2023gpt4}, the incorporation of language semantics to augment vision models has recently surfaced as a promising avenue in the realm of computer vision tasks \cite{radford2021learning, tsimpoukelli2021multimodal, alayrac2022flamingo, zhang2022glipv2}. Among these, Vision-and-Language Pre-training (VLP) \cite{mu2021slip, li2022grounded} has evolved into a popular paradigm across numerous vision-and-language tasks, owing to its ability to learn generalizable multimodal representations from extensive image-text data \cite{alayrac2022flamingo, chen2022align, Cheng_2022_CVPR}. These methods have recently found applications in multimodal retrieval \cite{dzabraev2021mdmmt}, vision-and-language navigation \cite{anderson2018vision}, among others \cite{cao2022searching}. Nevertheless, implementing effective prompt-based learning on base models, particularly extracting high-order relationships in intricate scenarios, remains crucial for the successful deployment of large models. Thus, investigating how to efficaciously extract specific information from large models is of utmost importance.

\textbf{HOI Vision-and-Language Modeling (HOI-VLM).} Despite the moderate success of previous HOI detectors \cite{zhang2021mining, zhang2022efficient, liu2022interactiveness}, these often perceive interactions as discrete labels, overlooking the rich semantic text information encapsulated in triplet labels. Of late, a handful of researchers \cite{zhong2021polysemy, iftekhar2022look, li2022improving, wang2022learning, yuan2022rlip, yuan2022detecting} have delved into HOI Vision-and-Language Modeling in an effort to further enhance HOI detection performance. Among these, \cite{zhong2021polysemy}, \cite{iftekhar2022look}, and \cite{yuan2022detecting} have primarily focused on incorporating language prior features into HOI recognition. Meanwhile, RLIP \cite{yuan2022rlip} and \cite{wang2022learning} have proposed the construction of a transferable HOI detector via the VLP approach. Serving as applications and extensions of Vision-and-Language learning within the HOI domain, these HOI-VLM methodologies strive to comprehend the content and relations between visual interaction features and their respective triplet texts. Nevertheless, these methodologies harness cross-modal knowledge in a markedly restricted manner, falling short of fully unleashing the potential of cross-modal knowledge and LLMs within the HOI domain.

\section{Method}
\subsection{Overview}
In traditional HOI detection task, the objective is typically to optimize a detector $\Phi_{\theta_\mathcal{D}}$, such as the most commonly-used Transformer-based detectors in recent years. By inputting a query $\mathcal{Q}^{ho}$ related to the Human-Object (HO) pairs, the HOI detector learns the location of the HO pairs and corresponding interactions in images. This optimization objective can be formulated as follows:
\begin{equation}
\min~\mathbb{E}_{(\I) \sim \X_\I} \left[ \mathcal{L} (\mathcal{GT},\Phi_{\theta_\mathcal{D}}(\I,\mathcal{Q}^{ho}))\right]
\end{equation}
where $\mathcal{GT}$ and $\mathcal{L}$ are ground-truth labels and overall loss function respectively, $\X_\I=\left\{\left(\I_i\right)\right\}_{i=1}^{|\mathcal{X}|}$ denotes the HOI image dataset, $\I$ represents the input image, and $\theta$ indicates the weights in HOI detector $\Phi_{\theta_\mathcal{D}}$. Thus, traditional HOI detectors rely on purely visual detection tasks for optimization.

In this work, our aim is to utilize superior VL foundation models and additional interaction interpretation $\mathcal{T}$ during the training and inference process. In this way, we can improve the optimization objective as follows:
\begin{equation}
\min~\mathbb{E}_{(\I, \mathcal{T}) \sim \X} \left[ \mathcal{L} (\mathcal{GT},\Phi_{\theta_\mathcal{D}}(\I,\mathcal{Q}^{ho}), \Phi_{\theta_\mathcal{F}}(\I, \mathcal{T},\mathcal{P}^{ho}))\right]
\end{equation}
where $\Phi_{\theta_\mathcal{F}}$ denotes the Vision-Language (VL) foundation models and $\mathcal{P}^{ho}$ represents the HOI-specific prompts for $\Phi_{\theta_\mathcal{F}}$. $\X=\left\{\left(\I_i, \mathcal{T}_i\right)\right\}_{i=1}^{|\mathcal{X}|}$ is an image-text corpus, where $\I_i$ denotes the $i$-th input image and $\mathcal{T}_i$ represents the phrase annotations (\emph{e.g.} ``Human ride bicycle'') in $\I_i$. In zero-shot setting, we also use knowledge retrieval to enrich the texts $\mathcal{T}$. Details will be described in Section \ref{sec:Knowledge_Retrieval}.

\begin{figure*}[t]
\vspace{-6pt}
  \centering
  \includegraphics[width=\linewidth]{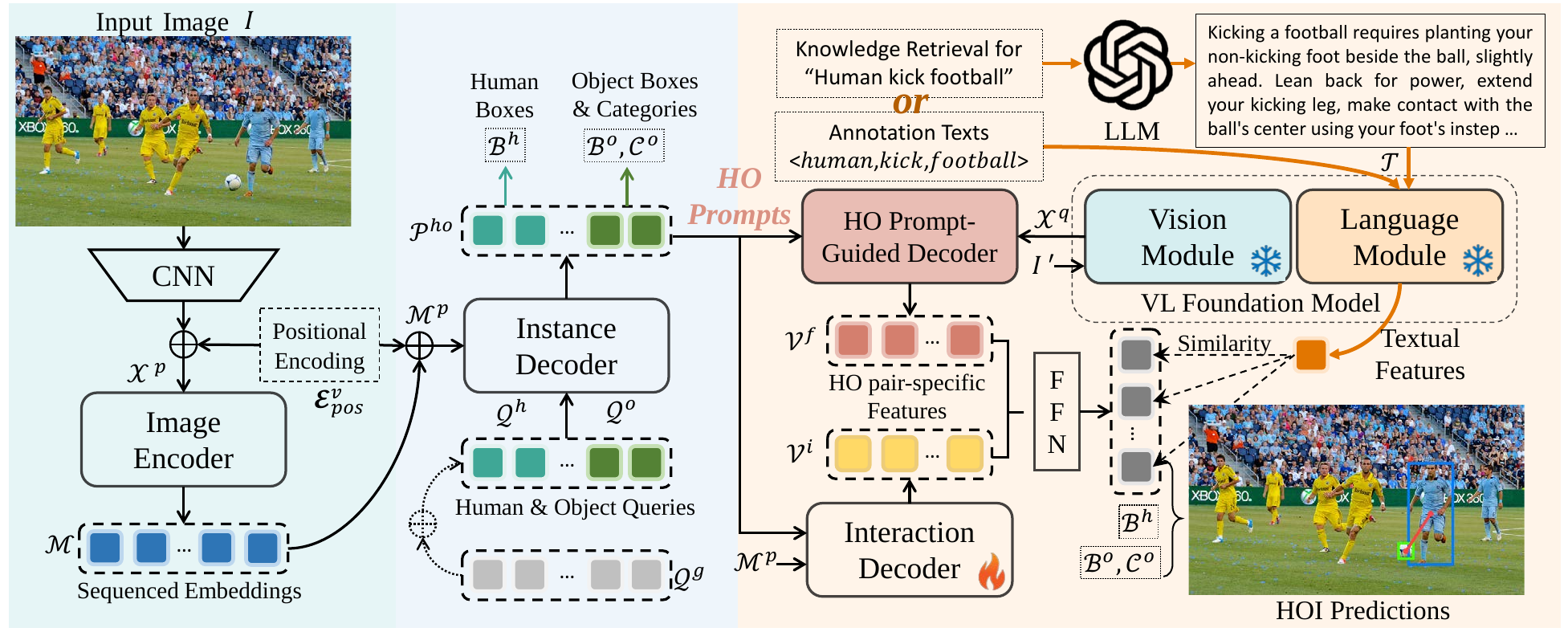}
   \caption{\textbf{Overview of the UniHOI.} Its feature hierarchy is divided into three levels. After the first two levels achieve instance detection for humans and objects, we designed a HO spatial prompt learning method, which extracts more generic higher-level relationship representations associated with HO pairs from the VL foundation models. For open-category interaction detection, UniHOI supports annotations or retrieved texts input, and then determine whether the interaction described in the input text occurs between each HO pair through feature similarity measurement.}
\vspace{6pt}
   \label{fig:2}
\vspace{-4mm}
\end{figure*}

In HOI detection, the detector not only needs to accurately locate the position of the interacting HO pair, but also needs to reliably model the interaction relationship occurring in the HO pair. But these two tasks have different requirements on the image feature level: the former focuses more on the edges and contours of instances, while the latter focuses more on high-level relational cues. To present and analyze this complicated process more clearly, we explicitly divide the feature learning in HOI detector into three stages: \emph{basic visual feature extraction}, \emph{instance-level feature learning}, and \emph{high-level relationship modeling}, as shown in Figure \ref{fig:2}. The basic visual feature extraction is usually performed by the backbone (\emph{e.g.} CNN). For the instance-level feature learning, we follow previous methodologies and design prompts related to HO locations to carry and transfer the position information, then apply the Hungarian algorithm for matching with the $\mathcal{GT}$, and compute the loss of the bounding box. The third stage, which is the key focus of this study, involves using spatial location prompts related to the HO pairs to conduct targeted prompt learning from the visual model. The details of each component are explained in the following sections.

\subsection{HO Spatial Prompts Generation}
We follow the Transformer-based detection architectures \cite{carion2020end} to conduct basic visual feature extraction and instance detection. Fed with an input image $\I \in \mathcal{R}^{H \times W \times C}$, the CNN generates a feature map $\X^v \in \mathcal{R}^{h \times w \times c}$. Then, $\X^v$ is compressed by a projection convolution layer with a kernel size $1 \times 1$. Next, a flatten operator is used to segment patch embeddings $\left\{x_1^v, x_2^v, \ldots, x_{N^v}^v\right\}$, where $N^v$ is the number of patch embeddings. The patch embeddings $\left\{x_1^v, x_2^v, \ldots, x_{N^v}^v\right\}$ is then linearly projected through a linear transformation $\mathcal{E}^v \in \mathcal{R}^{c \times D^v}$. Thus, the input for instance detection are calculated via summing up the projected embeddings and position embeddings $\mathcal{E}_{p o s}^v \in \R^{N^v \times D^v}$:
\begin{equation}
\X^p \in \R^{N^v \times D^v}=\left[ x_1^v \mathcal{E}^v ; x_2^v \mathcal{E}^v ; \ldots ; x_{N^v}^v \mathcal{E}^v\right]+\mathcal{E}_{p o s}^v
\end{equation}

After patching and sequencing the features extracted by CNN, we follow previous methods to perform self-attention on $\X^p$: $\mathcal{M} = \Phi_{\theta_\mathcal{IE}}(\X^p) \in \R^{N^v \times D^v}$. This process is implemented by a Transformer encoder $\Phi_{\theta_\mathcal{IE}}$, which is composed of $N$ Transformer encoder layers.

For instance-level feature learning, we deploy an instance decoder $\Phi_{\theta_\mathcal{ID}}$ to locate the positions of humans and objects. Based on the $\mathcal{M}$, humans and objects are detected through the human query set $\mathcal{Q}^{h} \in \R^{N^q \times D^v}$ and the object query set $\mathcal{Q}^{o} \in \R^{N^q \times D^v}$ individually. Additionally, a position guided embedding \cite{liao2022gen} $\mathcal{Q}^{g} \in \R^{N^q \times D^v}$ is added to assign the human and object queries at the same position as a pair. The calculation process of instance decoder $\Phi_{\theta_\mathcal{ID}}$ can be formulated simply as follows:
\begin{equation}
\mathcal{P}^{ho} \in \R^{2N^q \times D^v}  = \left[\mathcal{P}^{h},\mathcal{P}^{o} \right] = \Phi_{\theta_\mathcal{ID}}(\mathcal{M}, \left[ \mathcal{Q}^{h} + \mathcal{Q}^{g},\mathcal{Q}^{o} + \mathcal{Q}^{g} \right])
\end{equation}
where the $\mathcal{P}^{ho}$ is the spatial tokens highly correlated with HO pair location information, the $N^q$ is the number of queries. With instance detection head $\mathcal{FFN}s$, we can calculate the human bounding boxs $\B^h \in \R^{N^q \times 4}$ from $\mathcal{P}^{h}$, object bounding boxs $\B^o \in \R^{N^q \times 4}$ and object categories $\C^o \in \R^{N^q \times N^c}$ from $\mathcal{P}^{o}$:

\begin{equation}
\left[\mathcal{B}^{h};\mathcal{B}^{o};\mathcal{C}^{o} \right] = \mathcal{FFN}s(\left[\mathcal{P}^{h},\mathcal{P}^{o} \right])
\end{equation}
where the feature tokens $\left[\mathcal{P}^{h},\mathcal{P}^{o} \right]$ are excellent spatial position features for HO pairs. Driven by prompt learning methods, we explore using the HO position features here to perform prompt learning on the VL foundation models.

\subsection{Prompting Foundation Models for HOI Modeling}
\label{sec:Prompting}
Considering the excellent performance of BLIP2 \cite{li2023blip} in cross-modal tasks and the fact that its Transformer architecture also integrates well with our HOI framework, in this work we use BLIP2 as the VL foundation model to support the recognition process of HOI. However, accurately extracting HOI related features from VL foundation model is the key to successfully applying VL to HOI tasks.

To reduce the computational cost of the VL foundation model, we first downsample the input image $\I$ to $\I'$ with dimension $W' \times H'$, and then use an image encoder $\Phi_{\theta_\mathcal{I}}$ to convert downsampled image into a feature map $\mathcal{X}^{f} \in \R^{w' \times h' \times c'}$. We then employ Q-Former  $\Phi_{\theta_\mathcal{Q}}$ in BLIP2 \cite{li2023blip} to learn higher-level content $\mathcal{X}^{q} \in \R^{N^f \times D^f}$ in image:
\begin{equation}
\X^q=\Phi_{\theta_\mathcal{F}}(\I')=\Phi_{\theta_\mathcal{I}} \circ \Phi_{\theta_\mathcal{Q}} (\I,\mathcal{Q}^{f})
\end{equation}
where the $\mathcal{Q}^{f}$ denotes the query in  $\Phi_{\theta_\mathcal{I}}$. These modules in $\Phi_{\theta_\mathcal{I}}$ are more capable of understanding content in images than regular visiual detectors due to extensive training on a wider range of data.

Then, we consider how to extract the complex interaction representations between specific humans and objects in foundation model. To associate the higher-level information in foundation model with HO pairs, we design a HO Prompt-guided Decoder (HOPD) $\Phi_{\theta_\mathcal{P}}$ to correlate the position information of HO pairs with feature tokens in foundation model:
\begin{equation}
\mathcal{V}^f=\Phi_{\theta_\mathcal{P}}((\mathcal{P}^{h}+\mathcal{P}^{o})/2,\X^q)
\end{equation}
where the $\mathcal{V}^f$ denotes the high-level information tokens corresponding to HO pairs. In this manner, the HOI-specific tokens is generated under the guidance of the guidance of the human-object query pairs. This spatial prompts act as query embeddings in HOPD, interact with each other through self-attention layers, and interact with frozen image features through cross-attention layers (inserted every other transformer block in HOPD). In this way, these spatial prompts are fed into HOPD, which in turn goes on to query information about the relationships between their respective corresponding HO pairs. In addition, interaction decoding is performed in the visual HOI detector:
\begin{equation}
\mathcal{V}^i=\Phi_{\theta_\mathcal{IN}}((\mathcal{P}^{h}+\mathcal{P}^{o})/2,\M)
\end{equation}
Finally, we concatenate $\mathcal{V}^f$ and $\mathcal{V}^i$ as the visual representations of the HO pairs and feed them to a FFN to predict the interaction category. For model training, we adhere to the precedent set by query-based methods \cite{carion2020end}, utilizing the Hungarian algorithm to assign a corresponding prediction for each ground-truth instance. The loss and hyperparameters used for model optimization in this paper follow those of previous work \cite{liao2022gen}. More details are provided in the Appendix.


\subsection{Knowledge Retrieval for HOI Reasoning in Open World}
\label{sec:Knowledge_Retrieval}
In previous zero-shot HOI detectors, the recognition of unseen categories is usually achieved by directly using word embeddings and visual features for similarity measures. The problem with this approach is that it relies only on simple word embeddings, which may not be sufficiently detailed and rich in the representation of HOIs. The model may not recognize some special categories well enough. Inspired by the human understanding of interactions, we propose a method for HOI detection in an open world based on descriptive text.

The impressive performance of large-scale pretrained language models, as well as the potentially enormous amount of implicitly stored knowledge, raises extensive attention about using language models as an alternative to conventional structured knowledge bases (LMs-as-KBs). In particular, the rapid development of the chatGPT family of LLM models in the last two years inspired us to obtain richer and more comprehensive knowledge descriptions, called GPTs-as-KBs. Theoretically, the performance of the visual model can be improved theoretically by transferring a larger amount of knowledge that can be effectively integrated into the visual model. Given an input image $\I^i$ and corresponding descriptions $\mathcal{T}_i$ in $\X=\left\{\left(\I_i, \mathcal{T}_i\right)\right\}_{i=1}^{|\mathcal{X}|}$, different phrase descriptions are linked with different HO pairs. We propose expanding the existing phrase descriptions $\mathcal{T}^i$ to obtain richer and more comprehensive knowledge descriptions $\mathcal{K}^i$. These descriptive knowledge facilitate more accurate interaction descriptions in zero-shot and open-vocabulary HOI recognition.

To obtain richer and more comprehensive knowledge descriptions, we design a knowledge retrieval process for LLM. We have designed a knowledge query statement for chatGPT, such as ``\texttt{Knowledge retrieve for $verb\_objection$, limited to $N$ words}'', to guide LLMs in explaining the interaction categories we are interested in. In this way, the transferable textual knowledge will be richer and the interpretation of the interactions will be closer to the way humans understand them. As an illustration, let us consider the phrase "Human ride bicycle" and "Human throw frisbee" . The knowledge retrieval process yields the following example result:
\begin{itemize}

\item \emph{\textbf{Riding a bicycle} involves balance, coordination, and physical exertion. The rider mounts the bike, propels forward by pushing pedals with their feet. Steering is achieved by turning handlebars. Brakes slow or stop the bike. Helmets are worn for safety.}

\item \emph{\textbf{Throwing a frisbee} involves grasping the disc, typically with a forehand grip, then swinging the arm and releasing the frisbee at the right moment for it to glide through the air. Direction and distance depend on the angle and speed of the throw. It's a common recreational activity.}

\end{itemize}
These detailed text descriptions can better guide the learning of visual features in open-category recognition. A more detailed open-category inference process is provided in the Appendix.


\section{Experiments}
\subsection{Implementation Details}
We evaluate our model on two public benchmarks, HICO-Det \cite{chao2018learning} and V-COCO \cite{gupta2015visual}. We adopt the BLIP2 \cite{li2023blip} model as the VL foundation model in our UniHOI. We map the 768-dimensional features output by Q-Former in BLIP2 to 256 dimensions to adapt to the feature dimension of the visual HOI detector. The number of layers in the HO Prompt-guided Decoder is identical to that in the Interaction Decoder. The output features of these two decoders are concatenated to generate the final category prediction. The model's optimization method, learning rate, training epochs, weight decay, and loss weights, among other hyperparameters, are set in accordance with previous methods to ensure fair comparison. Aside from the ablation studies in Section \ref{sec:Ablation_Studies} where we employed knowledge retrieval to further enhance model performance, in all other experiments we continue to use annotated phrases for model learning and inference to ensure fair comparison with other methods. All the experiments conducted with a batchsize of 16 on 8 Tesla V100 GPUs. The details of datasets, evaluation metrics, model parameters, computational overhead, \emph{etc.} will be introduced in Appendix.

\begin{figure*}[t]
    \centering
    \subfigure{\label{fig:vcoco-s}\includegraphics[width=0.32\linewidth]{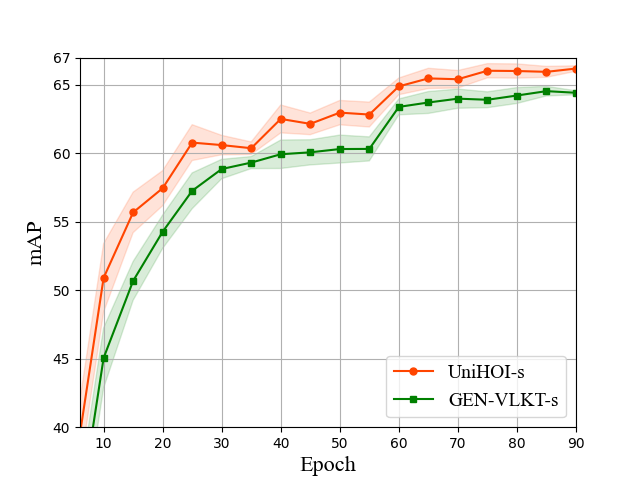}}
    \subfigure{\label{fig:vcoco-m}\includegraphics[width=0.32\linewidth]{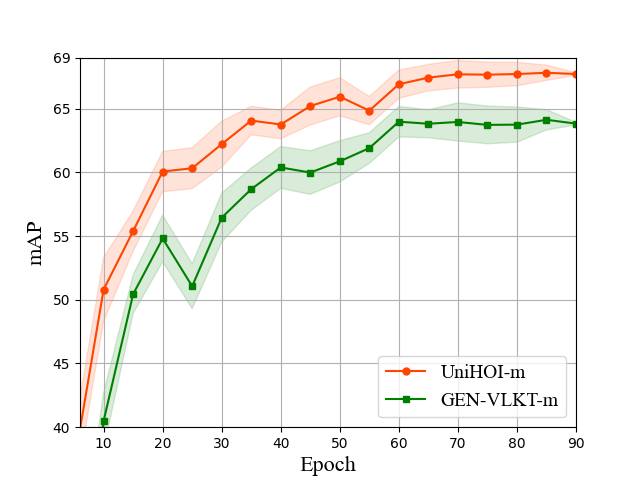}}
    \subfigure{\label{fig:vcoco-l}\includegraphics[width=0.32\linewidth]{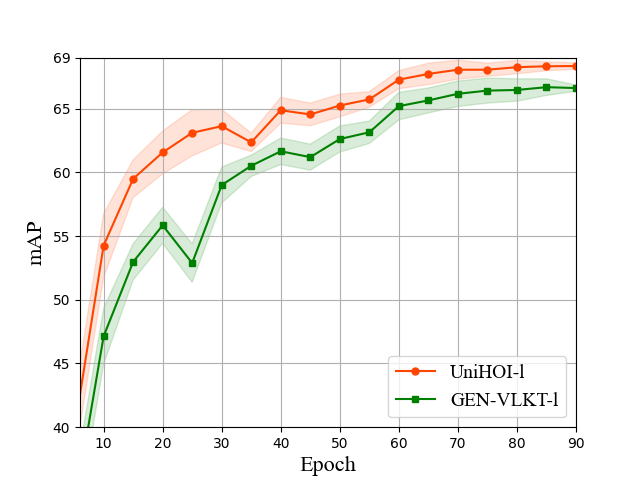}}
    \vspace{-10pt}
    \caption{\footnotesize The training process of \textbf{UniHOI} and previous State-of-the-Art \textbf{GEN-VLKT} \cite{liao2022gen} on V-COCO dataset.}
    \label{fig:training_process}
\vspace{-4mm}
\end{figure*}

\begin{table*}[t]
\caption{Experimental results on HICO-DET \cite{chao2018learning}. GEN-VLKT \cite{liao2022gen} is regarded as baseline.}
\centering
\footnotesize
\begin{tabular}{cccccccc}
\hline
            &            & \multicolumn{3}{c}{Default Setting} & \multicolumn{3}{c}{Known   Objects Setting}\\ \cline{3-8}
Method & Backbone   & Full       & Rare      & Non-rare     & Full         & Rare         & Non-rare       \\ \hline\hline
Two-stage Methods: \\
ATL \cite{hou2021affordance}                &ResNet-50      &23.81  &17.43  &27.42  &27.38  &22.09&28.96\\
VSGNet \cite{ulutan2020vsgnet}              &ResNet-152     &19.80  &16.05  &20.91  & -      & -& -\\
DJ-RN \cite{li2020detailed}                 &ResNet-50      &21.34  &18.53  &22.18  &23.69  &20.64&24.60\\
VCL \cite{hou2020visual}                    &ResNet-50      &23.63  &17.21  &25.55  &25.98  &19.12&28.03\\
DRG \cite{gao2020drg}                       &ResNet-50-FPN  &24.53  &19.47  &26.04  &27.98  &23.11&29.43\\
IDN \cite{li2020hoi}                        &ResNet-50      &24.58  &20.33  &25.86  &27.89  &23.64&29.16\\
FCL \cite{hou2021detecting}                 &ResNet-50      &25.27  &20.57  &26.67  &27.71  &22.34&28.93\\
SCG \cite{zhang2021spatially}               &ResNet-50-FPN  &29.26  &24.61  &30.65  &32.87  &27.89&34.35\\
UPT \cite{zhang2022efficient}               &ResNet-50      & 31.66  & 25.90  & 33.36   & 35.05  & 29.27  & 36.77 \\
UPT \cite{zhang2022efficient}               & ResNet-101    & 32.31 & 28.55  & 33.44 & 35.65 & 31.60 & 36.86 \\
ViPLO-s \cite{park2023viplo}           & ViT-B/32      & 34.95 & 33.83  & 35.28 & 38.15 & 36.77 & 38.56 \\
ViPLO-l \cite{park2023viplo}           & ViT-B/16      & 37.22 & 35.45  & 37.75 & 40.61 & 38.82 & 41.15 \\

\hline
One-stage Methods: \\
PPDM \cite{liao2020ppdm}                    &Hourglass-104  &21.94  &13.97  &24.32  &24.81  &17.09&27.12\\
HOI-Trans \cite{zou2021end}                 &ResNet-101     &26.61  &19.15  &28.84  &29.13  &20.98&31.57\\
AS-Net \cite{chen2021reformulating}         &ResNet-50      &28.87  &24.25  &30.25  &31.74  &27.07&33.14\\
QPIC \cite{tamura2021qpic}                  &ResNet-101     &29.90  &23.92  &31.69  &32.38  &26.06&34.27\\
SSRT \cite{iftekhar2022look}                &ResNet-101     & 31.34  & 24.31  & 33.32   & -  & -  & - \\
CDN-S \cite{zhang2021mining}                    &ResNet-50      & 31.44  & 27.39  & 32.64   & 34.09  & 29.63  & 35.42 \\
CDN-L \cite{zhang2021mining}           &ResNet-101     & 32.07  & 27.19  & 33.53   & 34.79  & 29.48  & 36.38  \\
Liu \emph{et al.} \cite{liu2022interactiveness}    &ResNet-50      & 33.51  & 30.30  & 34.46   & 36.28  & 33.16  & 37.21\\
MUREN \cite{kim2023relational}    &ResNet-50      & 32.87  & 28.67  & 34.12   & 35.52  & 30.88  & 36.91\\
GEN-VLKT-s \cite{liao2022gen}               &ResNet-50      & 33.75  & 29.25  & 35.10   & 36.78  & 32.75  & 37.99\\
GEN-VLKT-m \cite{liao2022gen}               &ResNet-101      & 34.78  & 31.50  & 35.77   & 38.07  & 34.94  & 39.01\\
GEN-VLKT-l \cite{liao2022gen}               &ResNet-101      & 34.96  & 31.18  & 36.08   & 38.22  & 34.36  & 39.37  \\
HOICLIP \cite{ning2023hoiclip}               &ResNet-101      & 34.69  & 31.12  & 35.74   & 37.61  & 34.47  & 38.54 \\
Xie \emph{et al.} (large) \cite{xie2023category}                &ResNet-101      & 36.03  & 33.16  & 36.89   & 38.82  & 35.51  & 39.81 \\
\hline
\rowcolor[HTML]{F0F0F0}
\textbf{\emph{UniHOI-s}} (w/ BLIP2)                & ResNet-50     & 40.06 & 39.91 & 40.11 & 42.20 & 42.60 & 42.08 \\
\rowcolor[HTML]{F0F0F0}
                                               &               & \red{(+6.31)} & \red{(+10.66)} & \red{(+5.01)} & \red{(+5.42)} & \red{(+9.85)} & \red{(+4.09)} \\
\rowcolor[HTML]{F0F0F0}
\textbf{\emph{UniHOI-m}} (w/ BLIP2)                & ResNet-101    & 40.74 & 40.03 & 40.95 & 42.96 & 42.86 & 42.98        \\
\rowcolor[HTML]{F0F0F0}
                                               &               & \red{(+5.96)} & \red{(+8.53)} & \red{(+5.18)} & \red{(+4.89)} & \red{(+7.92)} & \red{(+3.97)} \\
\rowcolor[HTML]{F0F0F0}
\textbf{\emph{UniHOI-l}} (w/ BLIP2)                & ResNet-101    & 40.95 & 40.27 & 41.32 & 43.26 & 43.12 & 43.25        \\
\rowcolor[HTML]{F0F0F0}
                                               &               & \red{(+5.99)} & \red{(+9.09)} & \red{(+5.24)} & \red{(+5.04)} & \red{(+8.76)} & \red{(+3.88)} \\
\hline
\end{tabular}
\label{tab1}
\vspace{-4mm}
\end{table*}

\begin{figure*}[htp]
  \centering
  \includegraphics[width=\linewidth]{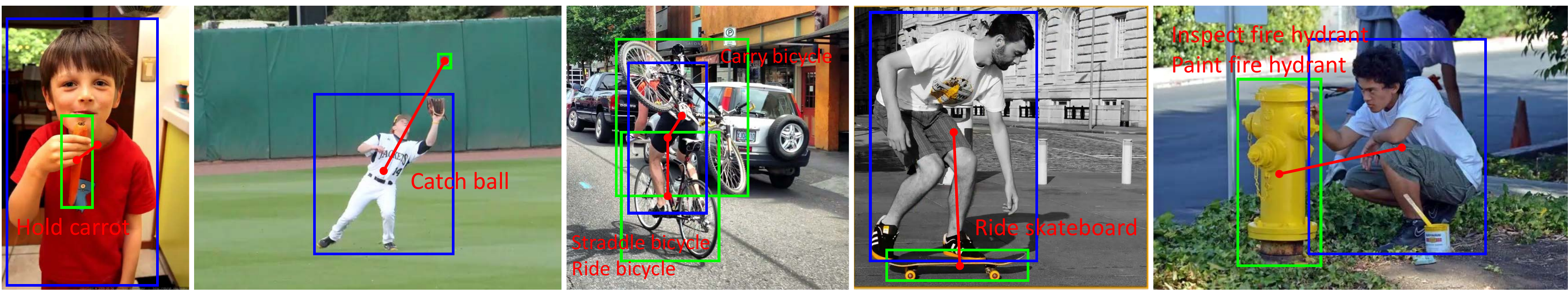}
   \caption{Detection results of our UniHOI in HICO-DET \cite{chao2018learning}.}
   \label{fig:4}
\vspace{-4mm}
\end{figure*}

\begin{table*}[t]
\caption{Comparison results with the methods using extra datasets on HICO-DET dataset \cite{chao2018learning}. For extra datasets, ``P'' indicates human pose and ``L'' indicates linguistic knowledge.}
\centering
\footnotesize
\setlength{\tabcolsep}{5pt}
\begin{tabular}{cccccccc|cc}
\hline
            &        & \multicolumn{6}{c|}{HICO-DET} & \multicolumn{2}{c}{V-COCO}  \\
            &        & \multicolumn{3}{c}{Default   Setting} & \multicolumn{3}{c|}{Known   Objects Setting} \\ \cline{3-8}
Method      & Extras & Full    & Rare   & Non-rare  & Full & Rare & Non-rare  &${AP}_{role}^{\# 1}$ & ${AP}_{role}^{\# 2}$  \\ \hline\hline
FCMNet \cite{liu2020amplifying}           & P+L  & 20.41      & 17.34     & 21.56    & 22.04   & 18.97   & 23.12  & 53.1 & -\\
PD-Net \cite{zhong2021polysemy}           & L  & 20.76  & 15.68  & 22.28   & 25.59 & 19.93   & 27.28    & -  & -\\
DRG \cite{gao2020drg}                     & P      & 24.53 & 19.47  & 26.04 & 27.98  & 23.11  & 29.43  & 51.0  & - \\
ConsNet-F \cite{liu2020consnet}          & P      & 24.39     & 17.10     & 26.56  & -  & -  & -   & 53.2  & -\\
RLIP-ParSeD \cite{yuan2022rlip}          & L      & 30.70     & 24.67     & 32.50  & -  & -  & -  \\
RLIP-ParSe \cite{yuan2022rlip}           & L      & 32.84     & 26.85     & 34.63  & -  & -  & -  & 61.9  & 64.2 \\
PhraseHOI \cite{li2022improving}         & L      & 30.03     & 23.48     & 31.99  &33.74  & 27.35  & 35.64   & -  & -\\
OCN (large) \cite{yuan2022detecting}             & L      & 31.43     & 25.80     & 33.11  & -  & -  & -  & -  & -\\
HOICLIP \cite{ning2023hoiclip}           & L      & 34.69     & 31.12     & 35.74  & 37.61  & 34.47  & 38.54   & 63.5  & 64.8\\

\hline
\rowcolor[HTML]{F0F0F0}
\textbf{\emph{UniHOI-s}} (w/ BLIP2)  & L    &  40.06 & 39.91 & 40.11 & 42.20 & 42.60 & 42.08  & {65.58} & {68.27}      \\
\rowcolor[HTML]{F0F0F0}
\textbf{\emph{UniHOI-m}} (w/ BLIP2)  & L    & 40.74 & 40.03 & 40.95 & 42.96 & 42.86 & 42.98  & {67.95} & {70.61}      \\
\rowcolor[HTML]{F0F0F0}
\textbf{\emph{UniHOI-l}} (w/ BLIP2)  & L    & 40.95 & 40.27 & 41.32 & 43.26 & 43.12 & 43.25  & {68.05} & {70.82}      \\
\hline
\end{tabular}
\label{tab4}
\vspace{-4mm}
\end{table*}

\begin{table}
\begin{minipage}{.5\linewidth}
\caption{Experimental results on V-COCO.}
\scriptsize
\setlength{\tabcolsep}{3pt}
\centering
\begin{tabular}{ccc}
\hline
Method &${AP}_{role}^{\# 1}$ & ${AP}_{role}^{\# 2}$   \\ \hline\hline
VSGNet \cite{ulutan2020vsgnet}              &51.8&57.0\\
IDN \cite{li2020hoi}                        &53.3&60.3\\
UPT \cite{zhang2022efficient}               & 60.7 & 66.2 \\
VIPLO-s \cite{park2023viplo}               & 60.9 & 66.6 \\
VIPLO-l \cite{park2023viplo}               & 62.2 & 68.0 \\

\hline
HOTR \cite{kim2021hotr}                     &55.2&64.4\\
QPIC \cite{tamura2021qpic}                  &58.8&61.0\\
CDN \cite{zhang2021mining}                & 63.91 & 65.89 \\
Liu \emph{et al.} \cite{liu2022interactiveness}    & 63.0 & 65.2 \\
GEN-VLKT-s \cite{liao2022gen}               & 62.41 & 64.46\\
GEN-VLKT-m \cite{liao2022gen}               & 63.28 & 65.58\\
GEN-VLKT-l \cite{liao2022gen}               & 63.58 & 65.93  \\
HOICLIP \cite{ning2023hoiclip}              & 63.50 & 64.80  \\
Xie \emph{et al.} (large) \cite{xie2023category} & 66.50 & 69.90  \\
\hline
\rowcolor[HTML]{F0F0F0}
\textbf{\emph{UniHOI-s}} (w/ BLIP2)             & \textbf{65.58} (\textcolor{red}{+3.17}) & \textbf{68.27} (\textcolor{red}{+3.81})        \\
\rowcolor[HTML]{F0F0F0}
\textbf{\emph{UniHOI-m}} (w/ BLIP2)            & \textbf{67.95} (\textcolor{red}{+4.67}) & \textbf{70.61} (\textcolor{red}{+5.03})        \\
\rowcolor[HTML]{F0F0F0}
\textbf{\emph{UniHOI-l}} (w/ BLIP2)            & \textbf{68.05} (\textcolor{red}{+4.47}) & \textbf{70.82} (\textcolor{red}{+4.89})        \\
\hline
\end{tabular}
\label{tab2}
\end{minipage}
\hspace{-0.5cm} 
\begin{minipage}{.5\linewidth}
\caption{Zero-shot results on HICO-DET \cite{chao2018learning}.}
\scriptsize
\setlength{\tabcolsep}{2pt}
\begin{tabular}{ccccc}
\hline
Method & Type   &Unseen &Seen  &Full   \\ \hline\hline
Shen \emph{et al.} \cite{shen2018scaling}  &UC  &5.62  &-  &6.26 \\
FG \cite{bansal2020detecting}  &UC  &10.93  &12.60  &12.26 \\
ATL \cite{hou2021affordance}   &UC  &16.99  &20.51  &19.81 \\
\hline
VCL \cite{hou2020visual}       &RF-UC  &10.06 &24.28 &21.43 \\ 
ATL \cite{hou2021affordance}   &RF-UC  &9.18  &24.67 &21.57 \\
FCL \cite{hou2021detecting}   &RF-UC  &13.16 &24.23 &22.01 \\
GEN-VLKT \cite{liao2022gen}   &RF-UC  &21.36 &32.91 &30.56 \\
\rowcolor[HTML]{F0F0F0}
\textbf{\emph{UniHOI-s}} (w/ BLIP2) &RF-UC   &\textbf{28.68} (\textcolor{red}{+7.32}) &\textbf{33.16} (\textcolor{red}{+0.25})  &\textbf{32.27} (\textcolor{red}{+1.71})  \\
\hline

VCL \cite{hou2020visual}       &NF-UC  &16.22 &18.52 &18.06 \\ 
ATL \cite{hou2021affordance}   &NF-UC  &18.25 &18.78 &18.67 \\
FCL \cite{hou2021detecting}    &NF-UC  &18.66 &19.55 &19.37 \\
GEN-VLKT \cite{liao2022gen}    &NF-UC  &25.05 &23.38 &23.71 \\
\rowcolor[HTML]{F0F0F0}
\textbf{\emph{UniHOI-s}} (w/ BLIP2) &NF-UC   &\textbf{28.45} (\textcolor{red}{+3.4}) &\textbf{32.63} (\textcolor{red}{+9.25})  &\textbf{31.79} (\textcolor{red}{+8.08})  \\
\hline
GEN-VLKT \cite{liao2022gen}    &UO  &10.51 &28.92 &25.63 \\
\rowcolor[HTML]{F0F0F0}
\textbf{\emph{UniHOI-s}} (w/ BLIP2) &UO   & \textbf{19.72} (\textcolor{red}{+9.21}) &\textbf{34.76} (\textcolor{red}{+5.84})  &\textbf{31.56} (\textcolor{red}{+5.93})  \\
\hline
GEN-VLKT \cite{liao2022gen}    &UV  &20.96 &30.23 &28.74 \\
\rowcolor[HTML]{F0F0F0}
\textbf{\emph{UniHOI-s}} (w/ BLIP2) &UV   & \textbf{26.05} (\textcolor{red}{+5.09}) &\textbf{36.78} (\textcolor{red}{+6.55})  &\textbf{34.68} (\textcolor{red}{+5.94})  \\
\hline
\end{tabular}
\label{tab3}
\end{minipage} 
\vspace{-4mm}
\end{table}

\begin{figure*}[t]
  \centering
  \includegraphics[width=\linewidth]{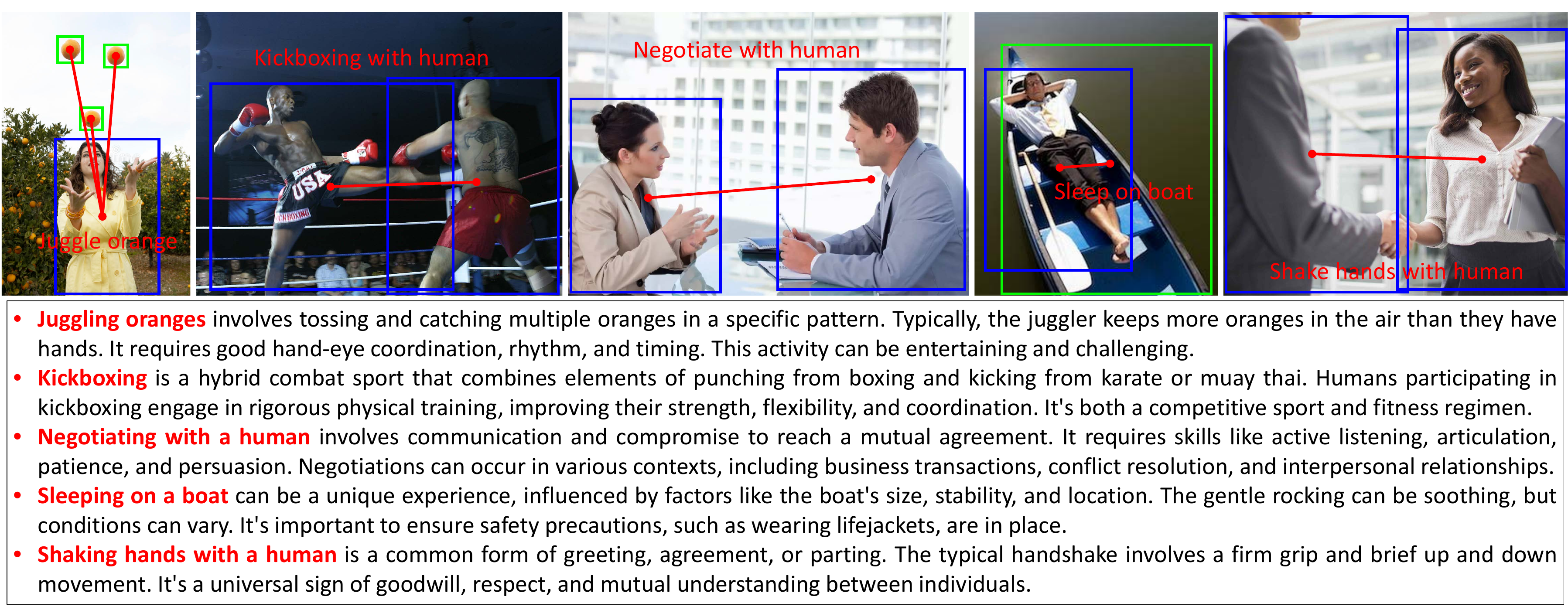}
\vspace{-4mm}
   \caption{HOI Detection in the Wild. UniHOI is capable of conducting open-category detection based on triplet phrases and also supports the input of descriptive texts retrieved by GPT \cite{openai2023gpt4}.}
   \label{fig:5}
\vspace{-3mm}
\end{figure*}

\begin{table}[t]
\begin{minipage}{.5\linewidth}
\caption{Ablation Studies on V-COCO \cite{gupta2015visual}.}
\footnotesize
\setlength{\tabcolsep}{5pt}
\centering
\begin{tabular}{lcc}
\hline
Method &${AP}_{role}^{\# 1}$ & ${AP}_{role}^{\# 2}$   \\ \hline
baseline                &62.41      &64.46\\
+ VL Foundation Model   &62.91      &64.83\\
+ HOPD                  &65.58      &68.27\\
+ Knowledge Retrieval   &\textbf{66.74}      &\textbf{69.31}\\
\hline
\end{tabular}
\label{tab:Ablation_Studies_vcoco}
\end{minipage}
\hspace{-0.5cm} 
\begin{minipage}{.5\linewidth}
\footnotesize
\caption{Ablation Studies on HICO-DET (UV).}
\setlength{\tabcolsep}{5pt}
\begin{tabular}{lcccc}
\hline
Method    &Unseen &Seen  &Full   \\ \hline
baseline                &20.96  &30.23  &28.74 \\
+ VL Foundation Model   &21.57  &31.62  &29.85 \\ 
+ HOPD                  &26.05  &36.78  &34.68 \\ 
+ Knowledge Retrieval   &\textbf{27.41}  &\textbf{37.82}  &\textbf{35.89} \\
\hline
\end{tabular}
\label{tab:Ablation_Studies_hico}
\end{minipage} 
\end{table}

\subsection{HOI Detection in the Closed World}
We compare the performance of our UniHOI with the previous representative and state-of-the-art methods such as ViPLO \cite{park2023viplo}, GEN-VLKT \cite{liao2022gen}, and HOICLIP \cite{ning2023hoiclip}. The detailed results of HICO-DET \cite{chao2018learning} and V-COCO \cite{gupta2015visual} are summarized in Table \ref{tab1} and \ref{tab2}, respectively. Generally, there are several observations drawn from diﬀerent aspects: (\emph{i}) The overall performance of our method is significantly superior to the previous state-of-the-art methods. (\emph{ii}) Whether the visual feature extractor adopts the ResNet-50 or ResNet-101, our method yields excellent performance, especially a stronger backbone that can bring further improvements. (\emph{iii}) As shown in Tables \ref{tab4}, our approach also has a significant performance lead compared to those methods that rely on additional data sets (\emph{i.e.}, Human Pose \cite{liu2020amplifying}, and Vision-and-Language \cite{yuan2022rlip}). Figure \ref{fig:training_process} shows the training process of UniHOI-s and GEN-VLKT-s on the V-COCO dataset. Figure \ref{fig:4} shows some prediction cases of UniHOI on HICO-DET.


\subsection{Comparisons with Methods that Utilize Extra Information}
Table \red{\ref{tab4}} presents several methods that utilize extra information (\emph{i.e.}, pose, and language) from additional datasets for HOI detection. Theoretically, the introduction of extra information can indeed further enhance visual models in specific domains. The results demonstrate that the proposed UniHOI still significantly outperforms previous methods.

\subsection{Effectiveness for Zero-Shot HOI Detection}
\textbf{Unseen Composition (UC).} As shown in Table \ref{tab3}, we follow the \cite{liao2022gen} to conduct three kind of zero-shot experiments on HICO-DET. We first evaluate UniHOI-s with UC setting. Compared with GEN-VLKT \cite{liao2022gen}, our UniHOI-s substantially outperforms previous state-of-the-art on both RF-UC and NF-UC. Among them, UniHOI are 7.32mAP higher than GEN-VLKT on Unseen of RF-UC, and 9.25mAP higher than GEN-VLKT on Seen of NF-UC. This improvement is mainly due to our efficient prompt learning method for the foundation model, so UniHOI can cope well with some unseen categories. In fact, GEN-VLKT also performs some knowledge transfer from CLIP, but it lags behind UniHOI in the effectiveness of knowledge transfer.

\textbf{Unseen Object (UO).} For UO setting, compared with GEN-VLKT, UniHOI improves Unseen, Seen, and Full by 9.21, 5.84, and 5.93, respectively. This result proves that our method can still understand the interaction between human and novel objects well.

\textbf{Unseen Verb (UV).} On the challenging UV setting, compared to GEN-VLKT, UniHOI improves by 9.21, 5.84 and 5.93 on Unseen, Seen, and Full, respectively. In fact, on the three zero-shot settings mentioned above, the HOI recognition capability of UniHOI have almost surpassed some supervised methods in Table \ref{tab1}. This once again attests to the superiority of our UniHOI for open-category HOI detection and its powerful capability to understand interactions.

\subsection{HOI Detection in the Wild}
In addition to the zero-shot experiments on the HICO dataset, we also conducted open-category testing on open-world. Our UniHOI method supports flexible text input for detecting corresponding interactions. UniHOI can not only recognize category texts directly (e.g., "Juggle orange") but also take a descriptive text input (which could be human-written or knowledge retrieved from GPT). Consequently, interactions in images are identified through text representation. We present these detection results, along with some GPT-generated description texts, in Figure \ref{fig:5}. Our test results indicate that open-vocabulary recognition based on interaction description texts is more reliable, and this phenomenon is also in line with our basic intuition. In the future, we will release our pretrained model and code to enable more extensive testing in the open world.

\subsection{Ablation Studies}
\label{sec:Ablation_Studies}
\textbf{Effect of Foundation Model.}
We conduct ablation studies on the new components within UniHOI (as shown in Tables \ref{tab:Ablation_Studies_vcoco} and \ref{tab:Ablation_Studies_hico}). The GEN-VLKT-s is regarded as the baseline, and then incorporated various components for testing on the fully supervised V-COCO dataset, as well as the HICO-DET dataset under a zero-shot setting. Initially, we simply integrated the BLIP2 into our method in a sequential manner. We then used learnable queries to extract interaction features from Q-Former and concatenate them with $\mathcal{V}^i$. However, the results did not yield a significant performance improvement.

\textbf{Impact of HO Spatial Prompting.} 
After realizing the alignment issue with the features of HO-pairs in the previous methods, we introduced the HO Prompt-guided Decoder, as presented in Section \ref{sec:Prompting}. The results clearly illustrate that the introduction of this component significantly improved all performance metrics, thereby validating the effectiveness of our prompting learning strategy.

\textbf{Effect of Knowledge Retrieval.} 
To substantiate the importance of utilizing LLM for knowledge retrieval in creating a more universal HOI detector, we also conduct supplementary experiments in this strategy. The results showed that the UniHOI model, when applying knowledge retrieval, achieved further performance improvements. We believe this approach opens new avenues for enhancement in the field of zero-shot HOI detection.

\textbf{Effect of Different Foundation Model.} 
To assess the impact of different Foundation Models on the performance of UniHOI, we also provide a comparative analysis between UniHOI equipped with CLIP and BLIP2. For CLIP, we feed its visual output representation into HOPD. However, given that CLIP's cross-modal contrastive training strategy leans more towards image-level representations \cite{radford2021learning}, its output dimensionality is considerably lower than that of BLIP2, resulting in a less detailed feature representation. Experimental results indicate that both the UniHOI versions with CLIP and BLIP2 as the foundational models showed performance enhancements, with the latter exhibiting a more pronounced improvement. These findings underscore the efficacy and generalization of our approach.

\begin{table}[t]
\caption{The results of UniHOI equipped with different foundation models (\emph{i.e.,} CLIP \cite{radford2021learning} and BLIP2 \cite{li2023blip}) on V-COCO \cite{gupta2015visual}.}
\setlength{\tabcolsep}{5pt}
\centering
\begin{tabular}{lcc}
\hline
Method &${AP}_{role}^{\# 1}$ & ${AP}_{role}^{\# 2}$   \\ \hline

GEN-VLKT-s \cite{liao2022gen}               & 62.41 & 64.46\\
GEN-VLKT-m \cite{liao2022gen}               & 63.28 & 65.58\\
GEN-VLKT-l \cite{liao2022gen}               & 63.58 & 65.93\\ \hline

\textbf{\emph{UniHOI-s}} (w/ CLIP)             & {62.92} (\textcolor{red}{+0.51}) & {65.67} (\textcolor{red}{+1.21})        \\
\textbf{\emph{UniHOI-m}} (w/ CLIP)            & {64.85} (\textcolor{red}{+1.57}) & {67.62} (\textcolor{red}{+2.04})        \\
\textbf{\emph{UniHOI-l}} (w/ CLIP)            & {64.93} (\textcolor{red}{+1.35}) & {67.86} (\textcolor{red}{+1.93})        \\ \hline

\textbf{\emph{UniHOI-s}} (w/ BLIP2)             & \textbf{65.58} (\textcolor{red}{+3.17}) & \textbf{68.27} (\textcolor{red}{+3.81})        \\
\textbf{\emph{UniHOI-m}} (w/ BLIP2)            & \textbf{67.95} (\textcolor{red}{+4.67}) & \textbf{70.61} (\textcolor{red}{+5.03})        \\
\textbf{\emph{UniHOI-l}} (w/ BLIP2)            & \textbf{68.05} (\textcolor{red}{+4.47}) & \textbf{70.82} (\textcolor{red}{+4.89})        \\
\hline
\end{tabular}
\vspace{-4mm}
\end{table}

\section{Conclusion}
In this work, we propose UniHOI, a universal HOI detection framework. By utilizing large foundation model, spatially prompting the large model, and generalizing the HOI detector to the open world. UniHOI supports open-category interaction relationship recognition from annotation texts or descriptions, which can be flexibly extended to applications in the open world. Extensive experiments on public datasets and diverse settings demonstrate its strong universality - it behaves the strongest open category interaction recognition ability so far. We believe our research will stimulate following research along the universal high-level relationship modeling direction in the future.

{\small
\bibliographystyle{splncs04}
\bibliography{egbib}
}

\appendix

\onecolumn


\setcounter{page}{13}

The supplementary materials are organized as follows. In Appendix \ref{sec:Architecture_Details}, we present more detailed descriptions of our UniHOI architecture. In Appendix \ref{sec:Motivations}, we discuss the motivations for our UniHOI. In Appendix \ref{sec:Differences_clip}, we provide an in-depth explanation of differences between our UniHOI and previous CLIP-based methods. In Appendix \ref{sec:VL_Foundation_Models}, we examine the effects of VL foundation models of different scales. In Appendix \ref{sec:Training_details}, we provide an detailed explanation of the training and hyperparameter setting. In Appendix \ref{sec:Open_category_Inference}, we present detailed description of the open-category inference. In Appendix \ref{sec:Computational_Overhead}, we present the computational and inference costs of different components in UniHOI. In Appendix \ref{sec:Datasets_Metrics}, we outline the datasets and evaluation metrics used in our experiments.

\section{Architecture Details}
\label{sec:Architecture_Details}
In this study, the BLIP-2 ViT-L $\mathrm{OPT}_{2.7 \mathrm{B}}$ \cite{li2023blip} is adopted as the default Vision-Language (VL) foundation model for all experiments. Specifically, a pre-trained ViT-L/14 is utilized as the image encoder $\Phi_{\theta_\mathcal{I}}$ in BLIP2, while the pre-trained OPT \cite{zhang2022opt} serves as the Language Language Model (LLM). An input image, downsampled to $224 \times 224 \times 3$, is fed into the ViT-L, transforming it into $257 \times 1408$ dimensional Image tokens $\mathcal{X}^{f}$. Subsequently, a lightweight Q-Former $\Phi_{\theta_\mathcal{Q}}$ leverages a set of learnable query vectors to extract visual features from $\mathcal{X}^{f}$. In BLIP2  \cite{li2023blip}, the Q-Former acts as the bridge between the visual and language modalities, thereby effectively extracting representations that align with the textual context, which in turn facilitates the possibility of open-set HOI detection. In UniHOI, the $32 \times 768$ dimensional output of the Q-Former is passed through a Fully Connected (FC) layer to map it to $32 \times 256$ dimensions, thereby aligning with the feature dimensions of the visual HOI detector.

Following the setup of GEN-VLKT \cite{liao2022gen}, we have implemented three versions of UniHOI: the small version (UniHOI-s), the middle version (UniHOI-m), and the large version (UniHOI-l). Specifically, UniHOI-s employs ResNet-50 as the backbone network, while the latter two, UniHOI-m and UniHOI-l, utilize ResNet-101 as their backbone. For both UniHOI-s and UniHOI-m, the number of layers in the three Decoders—Instance, Interaction, and HO (Human-Object) Prompt-guided Decoder—is set to 3. In contrast, for UniHOI-l, each of the three Decoders comprises 6 decoder layers.

\section{Motivations for Our UniHOI Framework}
\label{sec:Motivations}
1) HOI detectors solely relying on supervised learning are inherently constrained in terms of the variety of categories they can handle, thereby limiting their generalizability and scalability.

\begin{figure*}[t]
\vspace{-6pt}
  \centering
  \includegraphics[width=\linewidth]{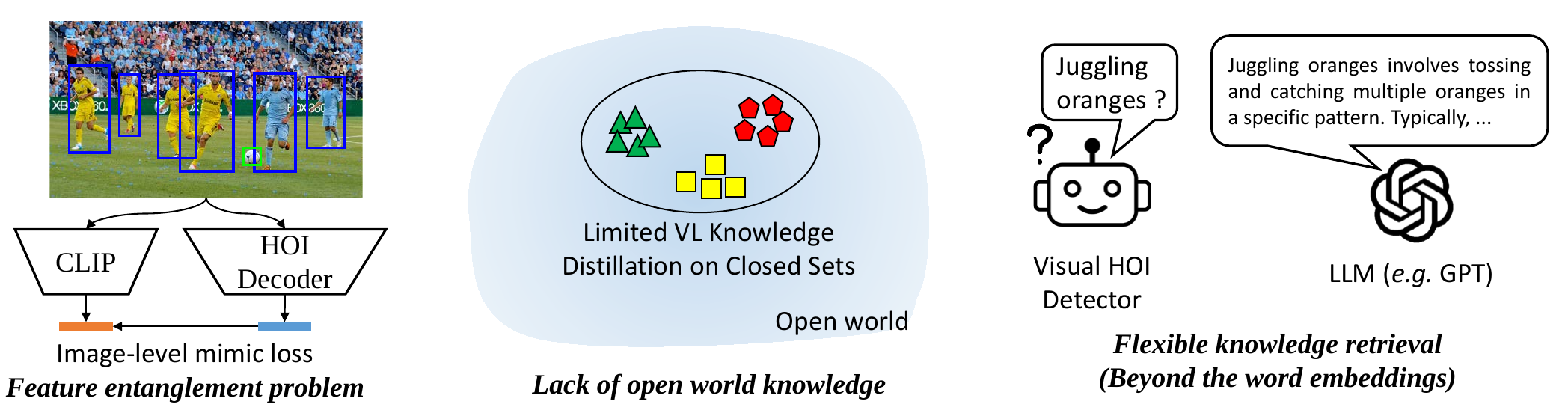}
   \caption{Our UniHOI focuses on the aforementioned issues in the current HOI detection field.}
\vspace{6pt}
   \label{fig:app1}
\vspace{-4mm}
\end{figure*}

2) While there have been text-driven zero-shot HOI detectors, they have not adequately addressed the cross-modal alignment issue in an open-world context. The underlying reasons are twofold: First, many methods are trained on a small amount of annotated HOI text, such as PhraseHOI \cite{li2022improving}, which limits the potential to extend HOI detectors to broader scenarios (see ``\textbf{\emph{Lack of open world knowledge}}'' in Figure \ref{fig:app1}). Second, these methods have not achieved HO pair-level cross-modal feature learning, predominantly seen in some CLIP-based methods \cite{radford2021learning}. This is due to CLIP's cross-modal contrastive learning approach that favors global representations over HO-pair level feature representation (see ``\textbf{\emph{Feature entanglement problem}}'' in Figure \ref{fig:app1}). We will elaborate on these details in the next section.

3) The paradigm of utilizing large VL foundation models to drive task-specific operations may become mainstream in the near future. This is because large VL foundation models encapsulate a more comprehensive world knowledge. If they are effectively applied to specific tasks, they can certainly propel advancements in more specialized fields. In the context of HOI detection, a crucial question is how to extract fine-grained higher-level human-object relationship information. How to effectively extract relationship features at the HO pair level from a VL foundation model represents a core issue that large model-driven HOI detection needs to address. In addition, richer descriptive knowledge could theoretically facilitate the understanding of interactions by visual HOI detectors (see ``\textbf{\emph{Flexible knowledge retrieval}}'' in Figure \ref{fig:app1}).

In summary, the aforementioned considerations collectively motivated our design for the UniHOI structure. The final experiments have confirmed the validity of these perspectives.

\section{Differences from the Previous CLIP-based Methods}
\label{sec:Differences_clip}
1) Previous CLIP-based methods exhibit limitations in the effectiveness of cross-modal learning. For instance, GEN-VLKT \cite{liao2022gen} performs visual mimic learning on the entire input image, which directly pools the features of the Interaction Decoder, using the mimic loss to learn the feature distribution of the CLIP Image Encoder. However, an input image may contain numerous human-object (HO) pairs, both interacting and non-interacting, leading to entangled features (see ``\textbf{\emph{Feature entanglement problem}}'' in Figure \ref{fig:app1}). Such a method of HOI knowledge distillation is inherently inefficient. Given CLIP's contrastive training strategy, it excels as a model adept at global representation. Therefore, our UniHOI attempts to exploit the knowledge of the VL foundation model from a more meticulous perspective of HO prompt learning.

2) CLIP-based methods' cross-modal learning didn't address the feature alignment problem of HO pairs. If the features at the HO level cannot be aligned, incorporating features from large models directly might lead to limited improvement. For example, HOICLIP \cite{ning2023hoiclip} tried to leverage CLIP's visual features and learn through cross-attention. However, its improvement compared to GEN-VLKT is still limited.

3) Many methodologies resort to data transfer on limited HOI datasets. For instance, PhraseHOI \cite{li2022improving} proposed an excellent idea for Joint Visual-and-Text Modeling. Nevertheless, HOI datasets can't cover the open world, restricting their adaptability to open categories.

4) Another common limitation of previous CLIP-based methods is the reliance on a naive text template. For instance, GEN-VLKT \cite{liao2022gen} and PhraseHOI \cite{li2022improving} both employ the template ``\texttt{A photo of a person [ACT] [OBJ]}''. In this case, the text knowledge that the model can leverage merely comprises two word embeddings, which in turn limits the effectiveness of knowledge transfer. Our UniHOI method can learn more intricate interaction relationships from more diverse interpretive statements (see ``\textbf{\emph{Flexible knowledge retrieval}}'' in Figure \ref{fig:app1}).

\section{Effects of VL Foundation Models with Different Scales}
\label{sec:VL_Foundation_Models}
We build UniHOI variants equipped with other VL foundation models and conduct comparison experiments on the V-COCO \cite{gupta2015visual} dataset to explore the effect of different large models. In Table \ref{VLFM}, we show the results of UniHOI-s equipped with different large models. These results indicate that different large models impact HOI recognition capability; generally, all these models promote our UniHOI framework to obtain state-of-the-art results on the V-COCO test set.

\begin{table*}[t]
\caption{Results on V-COCO with the UniHOI-s using different VL Foundation models.}
\centering
\begin{tabular}{ccc}
\hline
Methods &${AP}_{role}^{\# 1}$ & ${AP}_{role}^{\# 2}$   \\ \hline\hline
GEN-VLKT-s \cite{liao2022gen}               & 62.41 & 64.46\\
GEN-VLKT-m \cite{liao2022gen}               & 63.28 & 65.58\\
GEN-VLKT-l \cite{liao2022gen}               & 63.58 & 65.93  \\
\rowcolor[HTML]{F0F0F0}
\textbf{\emph{UniHOI-s}} (w/ BLIP-2 ViT-L $\mathrm{OPT}_{2.7 \mathrm{B}}$)  &65.58 &68.27   \\
\rowcolor[HTML]{F0F0F0}
\textbf{\emph{UniHOI-s}} (w/ BLIP-2 ViT-G $\mathrm{OPT}_{6.7 \mathrm{B}}$)  &\textbf{65.92} &68.56  \\
\rowcolor[HTML]{F0F0F0}
\textbf{\emph{UniHOI-s}} (w/ BLIP-2 ViT-L $\mathrm{FlanT5}_{\mathrm{XL}}$)  & 65.37 & 68.15  \\
\rowcolor[HTML]{F0F0F0}
\textbf{\emph{UniHOI-s}} (w/ BLIP-2 ViT-G $\mathrm{FlanT5}_{\mathrm{XXL}}$)  & 65.80 & \textbf{68.57}  \\
\hline
\end{tabular}
\label{VLFM}
\vspace{-4mm}
\end{table*}

\section{Training details and hyperparamters}
\label{sec:Training_details}
For model training, we follow the previous query-based methods \cite{carion2020end}, employing the Hungarian algorithm to assign a matching prediction for each ground-truth. The match incorporates both the boxes of human and object targets and the categories of interactions. The overall loss function consists of four components, namely box regression loss $\mathcal{L}_b$, the intersection-over-union loss $\mathcal{L}_u$, the classification loss $\mathcal{L}_c$ and the mimic loss $\mathcal{L}_m$:
\begin{equation}
\mathcal{L}=\lambda_b \sum_{i \in(h, o)} \mathcal{L}_b^i+\lambda_u \sum_{j \in(h, o)} \mathcal{L}_u^j+\sum_{k \in(o, a)} \lambda_c^k \mathcal{L}_c^k + \lambda_m \mathcal{L}_{m}
\end{equation}
where $\mathcal{L}_m$ represents an auxiliary loss employed for knowledge distillation from the language model to the visual HOI part. This approach has been proven effective for cross-modal learning in the \cite{liao2022gen}, so we retain this design in our work. We follow the GEN-VLKT \cite{liao2022gen} and QPIC \cite{tamura2021qpic} to set the $\lambda_b$, $\lambda_u$, $\lambda_c$ and $\lambda_m$ to 2.5, 1, 1 and 20.

\section{Open-category Inference}
\label{sec:Open_category_Inference}
In UniHOI, we harness the cross-modal alignment capabilities of the VL foundation model to perform vision-text content alignment. More specifically, we utilize the Image Encoder in BLIP-2 to extract image tokens, and then employ Q-Former to convert Image tokens into interaction representations that can be aligned with the features in text Encoder. The LLM is used to extract the descriptive text of the interaction relationship that needs to be recognized, for which we use the $[CLS]$ token as the text modality representation $\mathcal{T}^{cls}$. Subsequently, the cosine similarity ${Sim}(\cdot)$ is calculated between the visual modality representation and the text modality representation to determine whether the interaction described in the input text has occurred in the $j$-th HO pair:
\begin{equation}
{Logits}^{hoi}_j = Sim \left[ \mathcal{T}^{cls}, {FC}(Concat(\mathcal{V}^i_j,\mathcal{V}^{f}_j)) \right]
\end{equation}

\section{Computational Overhead}
\label{sec:Computational_Overhead}
Table \ref{tab:app_inference} showcases the computational overhead of UniHOI-s and GEN-VLKT methods of varying sizes when processing an $800\times600$ image, as well as their test results on the HICO-DET dataset. Here, GEN-VLKT-s, GEN-VLKT-m, and GEN-VLKT-l are structures proposed in \cite{liao2022gen}. To further investigate the HOI recognition performance of larger models, we extended the number of layers in the GEN-VLKT interaction decoder to 10 and 20 layers respectively, thus designing two variants, GEN-VLKT-xl and GEN-VLKT-xxl. Our results on the HICO-DET dataset suggest that continuing to enlarge the model following the original technical route of the visual HOI detector only yields limited improvement in HOI recognition performance. In comparison, our proposed UniHOI displays a significant accuracy advantage over larger visual HOI detectors, corroborating the superiority of our framework.

\begin{table*}[htp]
\caption{Test results of different methods on HICO-DET test set.}
\vspace{2mm}
\centering
\setlength{\tabcolsep}{2pt}
\begin{tabular}{cccccccc}
\hline
            &            & \multicolumn{3}{c}{Default Setting} & \multicolumn{3}{c}{Known   Objects Setting}\\ \cline{3-8}
Method & Inference Time (ms)   & Full       & Rare      & Non-rare     & Full         & Rare         & Non-rare       \\ 
\hline
GEN-VLKT-s   & 30.31    & 33.75  & 29.25  & 35.10   & 36.78  & 32.75  & 37.99\\
GEN-VLKT-m   & 34.92    & 34.78  & 31.50  & 35.77   & 38.07  & 34.94  & 39.01    \\
GEN-VLKT-l   & 45.62    & 34.96  & 31.18  & 36.08   & 38.22  & 34.36  & 39.37  \\
GEN-VLKT-xl  & 58.96    & 35.03  & 31.27  & 35.97   & 38.17  & 34.41  & 39.45  \\
GEN-VLKT-xxl & 91.57    & 35.23  & 31.42  & 36.17   & 38.39  & 34.62  & 39.68  \\
\rowcolor[HTML]{F0F0F0}
\textbf{\emph{UniHOI-s} (Ours)} & 85.86   & 40.06 & 39.91 & 40.11 & 42.20 & 42.60 & 42.08 \\
\rowcolor[HTML]{F0F0F0}
                                &     & \red{(+6.31)} & \red{(+10.66)} & \red{(+5.01)} & \red{(+5.42)} & \red{(+9.85)} & \red{(+4.09)} \\

\hline
\end{tabular}
\label{tab:app_inference}
\vspace{-4mm}
\end{table*}

\section{Datasets and Evaluation Metrics}
\label{sec:Datasets_Metrics}
\textbf{V-COCO} \cite{gupta2015visual}. V-COCO is a specialized derivative of the MS-COCO dataset, composed of 5,400 images allocated for training and 4,946 set aside for testing. This dataset encompasses a broad range of categories, specifically 80 object categories, 29 interaction categories, and an extensive total of 234 HOI categories. In adherence with previously established methodologies \cite{li2019transferable}, our evaluation focuses on a subset of 24 interaction classes. This choice is justified by the fact that four interaction classes lack object pairings, and one class suffers from a paucity of samples. For our evaluation metrics, we rely on the Mean Average Precision (mAP). For scenarios involving object occlusion, two different evaluation approaches are entertained. Scenario 1 adopts a stringent evaluation criterion necessitating a null bounding box prediction with coordinates [0, 0, 0, 0], while Scenario 2 provides a more lenient condition, disregarding the predicted bounding box in such occlusion instances for the purpose of evaluation.

\textbf{HICO-DET} \cite{chao2018learning}. HICO-DET is an expansive dataset designed for HOI detection, consisting of 37,536 images for training and an additional 9,515 for testing. It includes a considerable spectrum of categories, namely, 80 object categories, 117 interaction categories, and a substantial total of 600 HOI categories. Our evaluation approach aligns with prior methodologies \cite{li2019transferable}, focusing on the HICO-DET. We calculate the mAP metric under two distinct settings: Default and Known Objects. These settings are applied across three categories: Full (spanning all 600 HOI classes), Rare (consisting of 138 classes with fewer than 10 training samples), and Non-rare (comprising 462 classes with more than 10 training samples).

\end{document}